\documentclass[10pt,twocolumn,letterpaper]{article}

\usepackage{wacv}              

\usepackage{graphicx}
\usepackage{amsmath}
\usepackage{amssymb}
\usepackage{booktabs}
\usepackage{enumitem}
\usepackage{textcomp}
\usepackage{colortbl} 
\usepackage{xcolor}
\newcommand{\mat}[1]{\mathrm{\textbf{#1}}}
\newcommand{\vect}[1]{\mathrm{\textbf{#1}}}

\definecolor{mygreen}{rgb}{0, 0.8, 0}

\newcommand{\degree}{\textdegree\xspace}

\usepackage{array}
\usepackage{multirow}

\usepackage{tikz}

\usepackage[pagebackref,breaklinks,colorlinks]{hyperref}
\usepackage[export]{adjustbox}

\usepackage[capitalize]{cleveref}
\crefname{section}{Sec.}{Secs.}
\Crefname{section}{Section}{Sections}
\Crefname{table}{Table}{Tables}
\crefname{table}{Tab.}{Tabs.}

\def\wacvPaperID{898} 
\def\confName{WACV}
\def\confYear{2025}

\begin{document}

\title{\vspace{-0.25cm}GauFRe\hspace{0.1em}\includegraphics[width=1em]{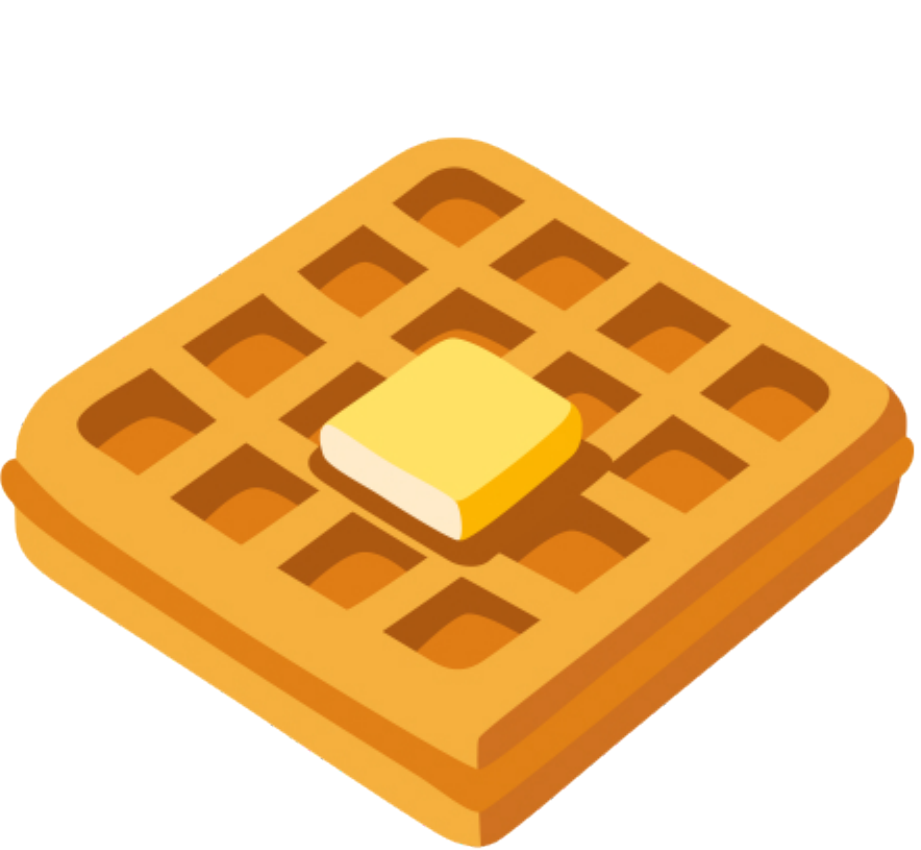}: Gaussian Deformation Fields for \\Real-time Dynamic Novel View Synthesis}
\author{
Yiqing Liang$^\ddagger$,
Numair Khan,
Zhengqin Li,
Thu Nguyen-Phuoc,\\
Douglas Lanman,
James Tompkin$^\ddagger$,
Lei Xiao
\\
Meta \hspace{1cm} $^\ddagger$Brown University
}

\twocolumn[{%
\renewcommand\twocolumn[1][]{#1}%
\maketitle
\begin{center}\vspace{-.7cm}
     \includegraphics[width=\textwidth, trim={0.0cm, 0.0cm, 0.08cm, 0cm}, clip]{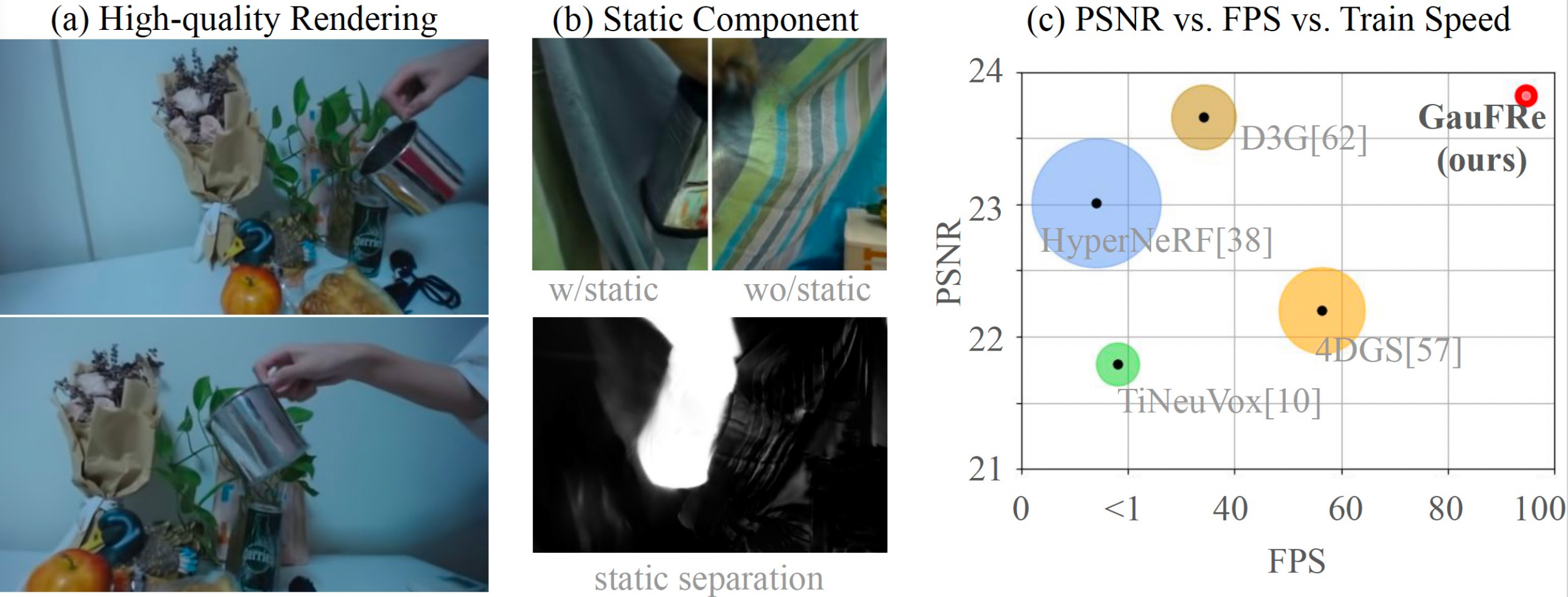}
     \vspace{-.6cm}
     \captionof{figure}{
     \textbf{(a)}: GauFRe's dynamic scene reconstruction results on the NeRF-DS~\cite{yan2023nerfds} real-world dataset. \textbf{(b)}: Our static component improves dynamic object rendering. 
     \textbf{(c)}: PSNR, real-time rendering performance, and optimization time (circle size) of state-of-the-art NeRF-based methods~\cite{TiNeuVox, park2021hypernerf} and Gaussian-Splatting-based methods~\cite{wu20234dgaussians, yang2023deformable} on NeRF-DS at 480$\times$270 resolution.
     }
     \label{fig:teaser}
 \end{center}
}]

\begin{abstract}
\vspace{-0.25cm}
  We propose a method that achieves state-of-the-art rendering quality and efficiency on monocular dynamic scene reconstruction using deformable 3D Gaussians. Implicit deformable representations commonly model motion with a canonical space and time-dependent backward-warping deformation field. Our method, GauFRe, uses a forward-warping deformation to explicitly model non-rigid transformations of scene geometry. 
  Specifically, we propose a template set of 3D Gaussians residing in a canonical space, and a time-dependent forward-warping deformation field to model dynamic objects. Additionally, we  tailor a 3D Gaussian-specific static component supported by an inductive bias-aware initialization approach which allows the deformation field to focus on moving scene regions, improving the rendering of complex real-world motion. The differentiable pipeline is optimized end-to-end with a self-supervised rendering loss.  Experiments show our method achieves competitive results and higher efficiency than both previous state-of-the-art NeRF and Gaussian-based methods. For real-world scenes, GauFRe can train in $\approx$20\ mins and offer 96\ FPS real-time rendering on an RTX 3090 GPU.
\end{abstract}

\vspace{-1cm}
\section{Introduction}
\label{sec:intro}
High-quality 3D reconstruction of dynamic scenes from RGB images is a persistent challenge in computer vision.
The challenge is especially great from monocular camera video: the setting is ill-posed as constraints on the surface geometry must be formed by simultaneously solving for an estimate of the scene's motion over time. 
Structure from motion~\cite{schoenberger2016mvs, schoenberger2016sfm} provides an estimate of rigid motion for static scenes, but real-world scenes have motions that extend beyond rigid or piecewise rigid to continual deformation, such as on human subjects. 
Given this challenge, one relaxation of the problem is to consider novel view synthesis instead, where we reconstruct the appearance of the scene to allow applications in editing to re-pose or re-time the scene.

For scenes with many cameras, optimization-based inverse graphics approaches can use image reconstruction losses to achieve high quality static or dynamic view synthesis. 
These often use neural networks as a function to predict the values of physical properties in a field, such as the density and radiance volumes within the influential neural radiance field (NeRF) technique \cite{mildenhall2020nerf}. 
Faster optimization and subsequent rendering can be achieved with the help of spatial indexing data structures, such as voxel grids~\cite{fridovich2022plenoxels}, octrees~\cite{yu2021plenoctrees}, and multi-scale hash tables~\cite{sun2021direct,muller2022instant}, or with proxy geometries such as planes~\cite{cao2023hexplane,fridovich2023k}.
Despite the spatial structures, these rely on computationally-expensive volume rendering to create an image, which requires many samples along each ray to render a pixel.

Following point-based graphics~\cite{zwicker2001surface,yifan2019differentiable}, differentiable primitive-based counterparts can be rasterized for faster speed. Representing a scene by a Gaussian primitive set is convenient as they are differentiable everywhere~\cite{Rhodin_2015_ICCV}, can be splatted in closed form~\cite{stoll2011fast,mejjati2021gaussigan}, and can be z-sorted efficiently under small-Gaussian assumptions without ray marching~\cite{khan2021diffdiffdepth}. 
Careful efficient implementation~\cite{kerbl2023gaussians,chen2024survey} leads to real-time static scene rendering at high resolutions, and overall produces competitive results.

Extending a primitive system to dynamic scenes is natural, with the idea that each Gaussian primitive represents a moving and deforming particle or blob/area of space tracked through time---the Lagrangian interpretation in the analogy to fluid flow. 
Directly optimizing each primitive's trajectory along time is easy in settings with sufficient constraints upon the motion of the flow, e.g., in 360\textdegree multi-camera settings~\cite{luiten2023dynamic}. 
For the under-constrained monocular video setting where constraints are sparse, it is challenging as accurate prediction of both geometry and motion are required, leading to failed reconstruction or low-quality output.

Backward-warping deformation fields (DFs) are well-studied for volume-rendering-based neural fields as a solution to the monocular challenge~\cite{Wang_2023_CVPR}. This approach samples multiple ray points at specific times and passes them into a backward-warping DF to query a canonical space of reconstruction~\cite{park2021hypernerf, park2021nerfies, pumarola2021d, Wang_2023_CVPR, TiNeuVox}. For primitive-based systems like Gaussian Splatting (GS), to render a ray, the status of all primitives at a timestep must be known at once, causing a backward-warping DF approach to be inefficient. 

Instead, our method uses a forward-warping DF design, where the DF predicts the deformed primitive system at a certain time. More specifically, GauFRe uses a Gaussian primitive set residing in a canonical space, and a forward-warping DF conditioned on time to estimate the temporal set.
We relate the forward-warping DF to how a physical 3D space point should change state along time \cite{luiten2023dynamic}, and parameterize the deformation for Gaussian attributes accordingly.  
Thanks to efficient GS optimization and real-time rendering with a CUDA rasterizer~\cite{kerbl2023gaussians}, GauFRe optimization takes $\approx$20~mins.\ instead of hours for a NeRF, with real-time rendering close to 100~FPS.

Beyond that, many regions of real-world dynamic scenes are static, and handling these separately can increase quality of both static and dynamic regions. Previous volume-rendering-based works have kept a static component parallel to the dynamic component, and combined the prediction from the two components for each queried ray sample. 
There is no ray sampling for primitive-based systems and so this design is no longer valid.
As such, we design a GS-specific static component with a separate non-deforming set of Gaussians that are initialized around SfM-derived 3D points~\cite{schoenberger2016sfm}, and combine this set with the deformable set to serve as the final primitive system.
Additionally, we show that an inductive-bias-aware Gaussian initialization helps to incorporate the static component.
The method, including the canonical GS primitive set, the non-deforming primitive set, and the forward-warping DF, is optimized end-to-end with self-supervised rendering loss.

In summary, we contribute:
\begin{enumerate}[itemsep=1pt,topsep=2pt]
    \item GauFRe, a forward-deformation-based dynamic scene representation of Gaussian primitives for real-time monocular-input dynamic view synthesis, enabling 96~FPS rendering on one RTX 3090.
    \item A GS-specific static component that represents static regions and further enhances reconstruction when complex motion is present.
    \item Experiments on both synthetic and real-world datasets show that GauFRe achieves competitive qualitative/quantitative results and efficiency compared with previous state-of-the-art methods.
\end{enumerate}
\section{Related Work}

Following the success of neural radiance fields (NeRFs)~\cite{mildenhall2020nerf} at static reconstruction, early  methods~\cite{park2021nerfies, pumarola2021d, li2020neural} proposed extending it to dynamic scenes by using an additional network to model motion. However, such methods must implicitly model both the 3D scene and its motion as a continuous representation using large MLPs. Thus they tend to be even more expensive to train and evaluate than the original, static radiance field representations. 

To address these shortcomings, recent approaches incorporate explicit constraints on the 3D space to train with simpler MLPs. 
Back to static reconstruction, variants of grid structure including planes are the most commonly used constraint~\cite{yu2021plenoctrees, sun2021direct, chen2022tensorf, fridovich2022plenoxels, karnewar2022relu}. Optimization-based point graphics~\cite{aliev2020neural, wiles2020synsin, xu2022point, zhang2022differentiable} are also popular. These include spherical proxy geometries~\cite{lassner2020pulsar}, splatting-based approaches~\cite{yifan2019differentiable, khan2021diffdiffdepth, kerbl2023gaussians}, methods for computing derivatives of points rendered to single pixels~\cite{ruckert2021adop}, and view-varying optimization of points~\cite{kopanas2021point}. 
Such explicit representations succeed in accelerating the optimization and rendering.%

Given above success, an intuitive approach is to extend them to dynamic scenes. 3D voxel grids can be extended into a fourth dimension for time. Unfortunately, the memory requirements of such a 4D grid quickly become prohibitive even for short sequences. As a result, a number of methods propose structures and techniques that reduce the memory complexity while still fundamentally being four-dimensional grids. 
Park~\etal~\cite{park2023temporal} extend Muller~\etal's multi-level spatial hash grid~\cite{muller2022instant} to 4D, and additionally allow for the separate learning of static and dynamic features. This latter capability allows the model to focus the representational power of the 4D hash grid on dynamic regions. Another approach factorizes the spatio-temporal grid into low-dimensional components. Jang and Kim~\cite{jang2022d} propose rank-one vectors or low-rank matrices, providing a 4D counterpart of the 3D tensorial radiance fields of Chen~\etal~\cite{chen2022tensorf}. Shao~\etal~\cite{shao2023tensor4d} hierarchically decompose the scene into three time-conditioned volumes, each represented by three orthogonal planes of feature vectors. An even more compact \textit{HexPlanes} solution by Cao and Johnson~\cite{cao2023hexplane} uses six planes, each spanning two of the four spatio-temporal axes. A similar decomposition is presented by Fridovich-Keil~\etal~\cite{fridovich2023k} as part of a general representation that uses ${d \choose 2}$ planes to factorize any arbitrary $d$-dimensional space. GauFRe models dynamic scene using deformable 3D-GS, achieving superior rendering quality, training speed and real-time rendering on real-world data in comparison.



\begin{figure*}[t!]
  \centering
   \includegraphics[width=0.9\linewidth, trim={0.8cm, 10cm, 18.8cm, 0cm}, clip]{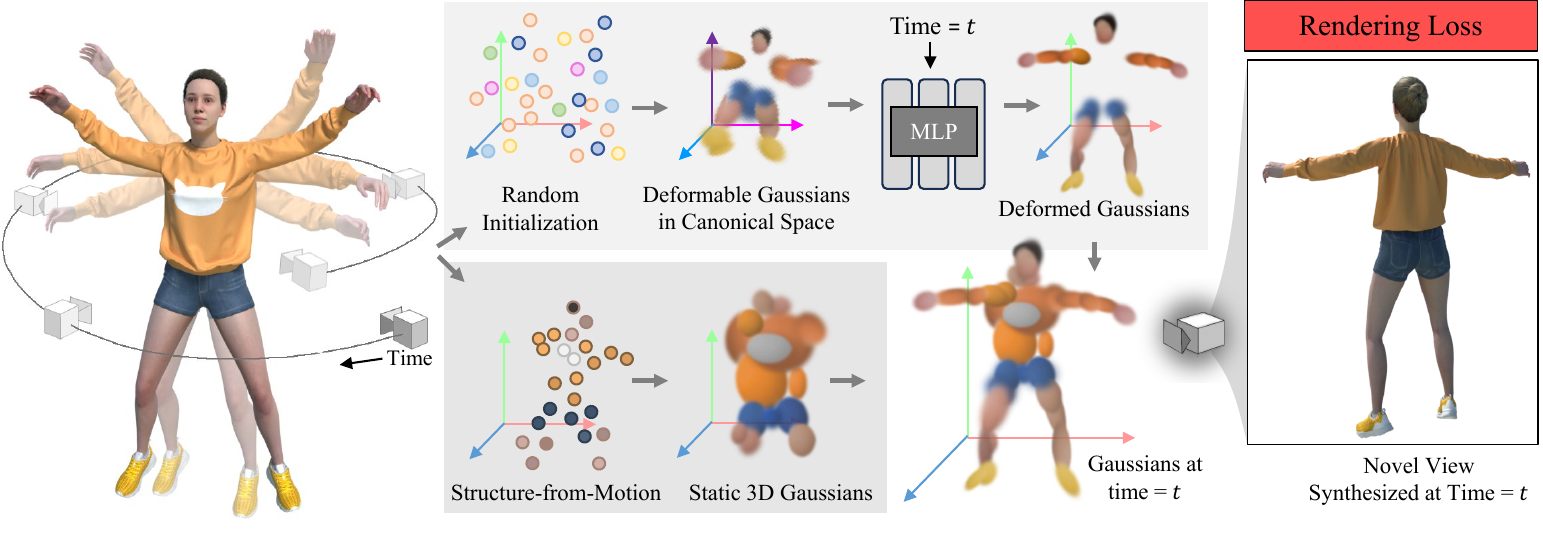}
   
   \vspace{-3mm}
   \caption{\textbf{An overview of our dynamic scene representation.} At each time frame $t$, our method reconstructs the scene as a combination of static and deformable anisotropic 3D Gaussians. The features of the deformable Gaussians are optimized in a canonical space and warped into frame $t$ using a deformation field. The static Gaussians are optimized in world space. 
   \vspace{-3mm}
   }
   \label{fig:overview}
\end{figure*}
Some NeRF-based works also attempt to split static and dynamic parts~\cite{li2020neural,wu2022d,Liang_2023_ICCV, liu2023robust,nerfplayer,wang2023masked,wang2022mixed,attal2023hyperreel}. They either make heuristic-based assumptions\cite{xian2021space,nerfplayer,wang2022mixed,wang2023masked}, or needs extra supervision than monocular RGB video\cite{Yoon_2020_CVPR,li2020neural,Gao-ICCV-DynNeRF, Liang_2023_ICCV,li2023dynibar,zhao2024pgdvs,liu2023robust,wang2022mixed,attal2023hyperreel}.  In comparison, GauFRe proposes a static component compatible with GS-based representation, does not make such assumption and only requires color image as input. \cite{wu2022d} has closest setting, but we enjoy quick training (mins vs. days) and real-time rendering.

\vspace{-0.25cm}
\paragraph{Dynamic Gaussian Splatting.} A number of papers propose using dynamic Gaussian representations\cite{duan20244d, katsumata2023efficient, yang2023gs4d, kratimenos2024dynmf, li2023spacetime, lin2023gaussian,
sun20243dgstream, Duisterhof2023MDSplattingLM,das2023neuralparametric, yu2023cogs,guo2024motionaware, lu2024gagaussian, liu2024modgs,icml2024-sp-gs}. Liuten~\etal~\cite{luiten2023dynamic} consider the 360\degree multi-camera case that is constrained in spacetime and take a Lagrangian tracking approach. Yang et al.~\cite{yang2023realtime} directly optimize 4D Gaussian position and color over time. Zielonka~\etal~\cite{Zielonka2023Drivable3D} use Gaussians to approach the problem of tracking driveable human avatars from multi-camera capture.   Similar to our approach, Wu~\etal~\cite{wu20234dgaussians} and Yang~\etal~\cite{yang2023deformable} both use a 
deformation field to model motion. 
Unlike our approach, neither considers static modeling.
 Further, we perform more detailed analysis of the design choices in a forward-warping DF approach, such that GauFRe sits within a sweet spot of competitive rendering quality with faster training and rendering.


\section{Method}


\subsection{Preliminaries: 3D Gaussian Splatting}
\label{sec:gaussian-splatting}
We use Kerbl~\etal's~\cite{kerbl2023gaussians} 3D Gaussians as our underlying scene representation. We recapitulate the primary aspects of their method to establish context for our discussion. 

A 3D scene is represented as a set of $n$ points $\{ \vect{x}_i \in \mathbb{R}^3, i=1, ...,n\}$. Each point is associated with features $(\mathbf{\Sigma}_i, \sigma_i, \vect{c}_i)$. These define the local radiance field as an anisotropic Gaussian distribution centered at $\vect{x}_i$ with covariance $\mathbf{\Sigma}_i \in \mathbb{R}^{3\times3}$, scalar density $\sigma_i$, and view-dependent color $\vect{c}_i$ represented by $m$-order spherical harmonics. Given a set of multi-view images of the scene and a suitable volumetric renderer, we can optimize a reconstruction objective over the set of Gaussians $\{ \mathcal{G}_i = (\vect{x}_i, \mathbf{\Sigma}_i, \sigma_i, \vect{c}_i) \}$ to represent the scene's global radiance field. To constrain $\mathbf{\Sigma}_i$ to a valid positive semi-definite covariance matrix during optimization, it is factored into a rotation matrix $\mat{R}_i\in \mathbb{R}^{3\times3}$ and scaling vector $\vect{s}_i \in \mathbb{R}^3$: 
\begin{equation}
\begin{array}{l}
        \mathbf{\Sigma}_i= \mat{R}_i \mathrm{Exp}(\vect{s}_i) {\mathrm{Exp}(\vect{s}_i}^T) {\mat{R}_i}^T \,
  \end{array}
  \label{eq:Sigma}
\end{equation} 
with the exponential activation preventing negative values while retaining differentiability. In practice, $\mat{R}_i$ is inferred from a unit-length quaternion $\vect{q}_i \in \mathbb{R}^4$ that provides better convergence behavior

The initial position $\vect{x}_i$ of the Gaussians is provided by a 3D point cloud obtained with a SfM (structure-from-motion) algorithm or randomly initialized. As the optimization proceeds, the Gaussians are periodically cloned, split, and pruned to achieve a suitable trade-off between rendering quality and computational resources.



Additionally, Kerbl~\etal demonstrate how the many continuous Gaussian radiance distributions can be efficiently rendered on graphics hardware. Given a target camera view transformation $\mat{V}$ and projection matrix $\mat{K}$, each $\mathcal{G}_i$ is reduced to a Gaussian distribution in screen space with projected mean $\vect{u}_i = \mat{K}\mat{V}\vect{x}_i \in \mathbb{R}^2$ and 2D covariance defined by the Jacobian $\mat{J}$ of $\mat{K}$ as
\begin{equation}
    \mathbf{\Sigma}_i'= \mat{J} \mat{V} \mathbf{\Sigma}_i \mat{V}^T \mat{J}^T
\end{equation}
%

\noindent
The 2D Gaussians are rasterized using elliptical weighted average splatting~\cite{zwicker2001surface}.

\subsection{Forward-warping Deformation Field}

\label{sec:gaussian-deformation-fields}
We model dynamics by augmenting Kerbl~\etal's representation with a time-conditioned deformation field. For volume-rendered representations such as NeRFs, deformation is commonly modeled as a backward warp from 4D spatio-temporal coordinates into a canonical 3D space~\cite{mildenhall2020nerf, park2021nerfies, pumarola2021d, Wang_2023_CVPR}. This is similar to a query search based on 3D position $\vect{x}$ and time $t$.
However, since 3D Gaussians represent the scene as an explicit set of primitives, we model the deformation as a forward warp instead. That is, the set of Gaussians $\{\mathcal{G}_i\}$ resides in the canonical space and a forward-warping field $\mathbf{\Phi}(\cdot)$ outputs the deformed set $\{\mathcal{G}_i^t\}$ with corresponding features $\{ (\vect{x}_i^t, \mathbf{\Sigma}_i^t, \sigma_i^t, \vect{c}_i^t)\}$, for each time step $t$. Thus, instead of a query search, the field explicitly represents changes in scene attributes over time. 





Among these attributes, we model rigid motion by deforming the position $\vect{x}_i$ and rotation $\vect{q}_i$. To allow the Gaussians to stretch and squeeze for capturing non-rigid transformations, we additionally deform $\vect{s}_i$. As a result, $\mathbf{\Phi}: \mathbb{R}^3\times\mathbb{R}^4\times\mathbb{R}^3 \rightarrow \mathbb{R}^3\times\mathbb{R}^4\times\mathbb{R}^3$. In comparison, the work of Luiten~\etal~\cite{luiten2023dynamic} only models rigid deformations. This relies on a dense multiview capture setting to allow accurate initialization of the Gaussian points from SfM, which makes their reconstruction problem easier than the monocular setting. While Katsuma~\etal~\cite{katsumata2023efficient} extend Luiten~\etal's approach to monocular settings, they freeze scaling during deformation which hurts reconstruction quality~(Fig.~\ref{fig:NeRF-DS-quali-concurrent}). 

Our deformation field models the $\delta$-change in the canonical space configuration of each attribute at time $t$:
\begin{equation}
\mathbf{\Phi}(\vect{x}_i, \vect{q}_i, \vect{s}_i, t) = (\delta\vect{x}_i^t, \delta\vect{q}_i^t, \delta\vect{s}_i^t)
\end{equation}
Then, the deformed attributes at time $t$ are given as,
\begin{align}
\vect{x}_i^t &= \vect{x}_i + \delta\vect{x}_i^t \\
\mathrm{Exp}(\vect{s}_i^t) &= \mathrm{Exp}(\vect{s}_i + \delta\vect{s}_i^t) \\
\vect{q}_i^t &= \vect{q}_i \cdot \delta\vect{q}_i^t
\end{align}
Note that the shape deformation $\delta\vect{s}_i^t$ can alternatively be a post-exponentiation delta: $\mathrm{Exp}(\vect{s}_i^t) = \mathrm{Exp}(\vect{s}_i) + \delta\vect{s}_i^t$. However, pre-exponentiation is a log-linear estimation problem which is simpler than an exponential one, and allows the optimization to handle negative changes. Practically, pre-exponentiation improves reconstruction quality (Table~\ref{tab:ablations}).

We multiply the rotation deformation $\delta\vect{q}_i^t$ with the canonical space value as combining two rotation operations mathematically corresponds to quaternion multiplication, rather than addition. To stabilize training, we normalize the quaternion after deformation. While addition is faster and works via the half-angle rotation, the reconstruction had more artifacts at novel viewpoints and timesteps as the resulting quaternion is not bound geometrically.

\vspace{-0.25cm}
\paragraph{Should We Deform Opacity and Color?}
In the presence of atmospheric effects, or when a scene entity changes color across time, the opacity $\sigma_i$ and appearance $\vect{c}_i$ of the canonical Gaussians should also deform. However, we found that deforming these attributes caused severe overfitting on training sets, and introduced more artifacts than were fixed in novel views. Thus, we choose not to deform  $\sigma_i$ and $\vect{c}_i$ over time and optimize them in the canonical space only.

\subsection{Static Component for Gaussian Splatting}
\label{sec:gaussian-static}


\begin{figure}[t!]
    \centering
    \includegraphics[width=\linewidth]{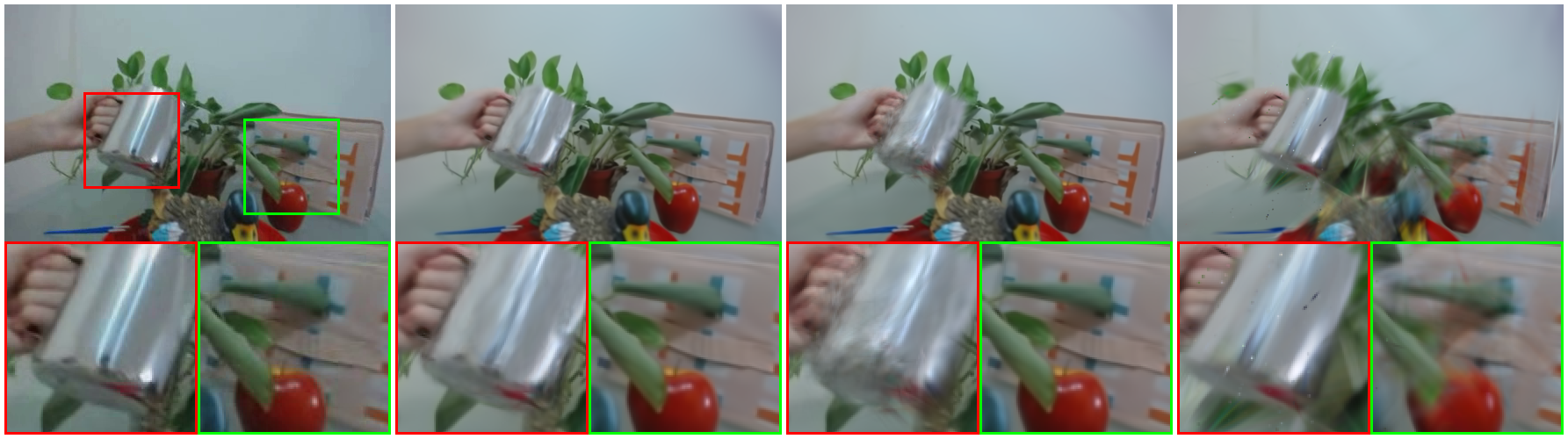}
    \scalebox{0.92}{\begin{tabular}{p{0.22\linewidth}p{0.22\linewidth}p{0.22\linewidth}p{0.22\linewidth}}
    \centering GT & \centering Ours & \centering Inseparate &  \centering 4DGS 
    \end{tabular}
    }
    \vspace{-5mm}
    \caption{\textbf{Separate deformable and static regions improves quality in dynamic regions.} \emph{Left to right:} Ground truth, our method, deformable Gaussians with no static component,  and the results of Wu et al.'s 4D Gaussians method~\cite{wu20234dgaussians}. }
    \label{fig:enable-static}
    \vspace{-0.25cm}
\end{figure}
In scenes with many static regions, we observe that the forward-warping deformation field sometimes struggles to represent motion adequately. Even though static scene regions do not need to change with time, a fully-deformable model will still spend capacity to describe tiny deformations in irrelevant regions due to noise in the camera pose or on the sensor. This issue is compounded by the fact that deformable Gaussians can easily densify to represent noise, dragging down efficiency. Therefore, if these regions can be ignored, then the network can focus its representational power and increase overall reconstruction quality.




To achieve this, we introduce a set of $k$ static  points $\{\vect{x}_j \in \mathbb{R}^3, j = 1, ..., k\}$ along with Gaussian features $\{ \mathcal{G}_j \}$ which resides alongside the deformable set $\{ \mathcal{G}_i\}$ (in a slight abuse of notation, we use the indices $i$ and $j$ to distinguish elements of the two sets). To render the scene at time $t$, we compute the deformed set $\{ \mathcal{G}_i^t \}$ and concatenate it to $\{\mathcal{G}_j\}$. The combined set is treated as a static world representation and rendered together. During optimization, the two sets are densified and pruned separately.



\begin{figure}[t!]
    \centering
       \includegraphics[width=0.24\linewidth]{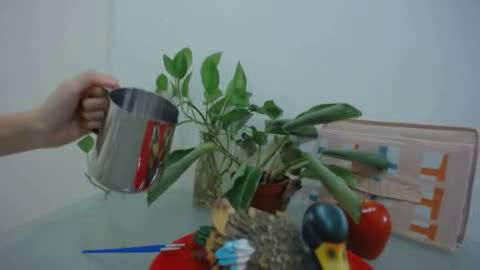}
       \includegraphics[width=0.24\linewidth]{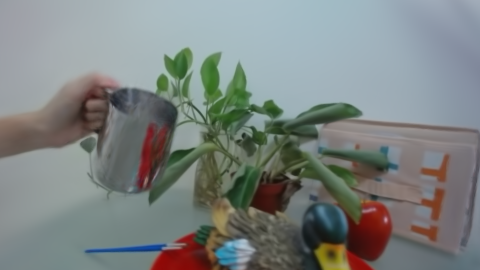}
       \includegraphics[width=0.24\linewidth]{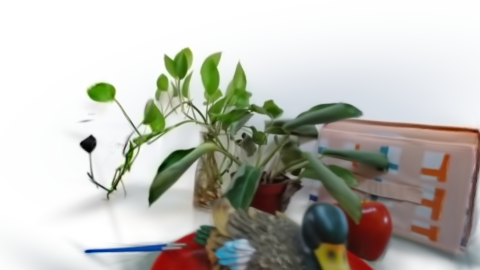}
       \includegraphics[width=0.24\linewidth]{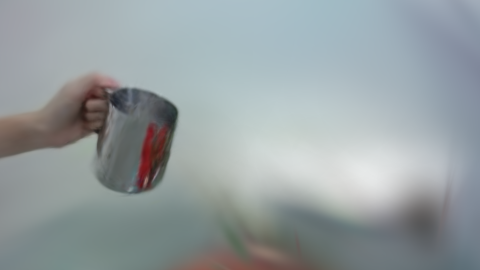}
       %
       \centering\scalebox{0.9}{\begin{tabular}{p{0.22\linewidth}p{0.22\linewidth}p{0.22\linewidth}p{0.22\linewidth}}
        \centering GT & \centering Full  & \centering Static  &  \centering Deformable  
        \end{tabular}}
    \vspace{-3mm}
    \caption{Visualizing the static and deformable 3D Gaussians optimized by our method. 
    }
    \vspace{-5mm}
\label{fig:static}
\end{figure}

\paragraph{Inductive Bias-Aware Initialization:} Kerbl~\etal~initialize their Gaussian primitives either with a pre-computed 3D point cloud from an SfM algorithm, or by uniformly sampling points within the scene bounding box. In our case, when initializing the 3D points $\{\vect{x}_i\}$ and $\{\vect{x}_j\}$ in the separate static and deformable sets, we found that if both are initialized with the sparse SfM point cloud, or both with uniform samples, the two sets fight each other over scene occupancy, leading to worse reconstruction than  the all-deformable case. To address this issue, we benefit from knowing that the feature-matching nature of SfM means the pre-computed point cloud only captures static scene regions. If the static set $\{ \mathcal{G}_j \}$ is initialized with SfM, it can quickly converge to its optimal state. On the other hand, initializing $\{\mat{G}_i\} $ with SfM points will take a long time, if ever, for the deformable set to converge. As such, we randomly initialize $\{\mathcal{G}_i\}$ with uniform samples, and initialize $\{\mathcal{G}_j\}$ with the SfM point cloud. As a result, the static Gaussian set is able to lighten the reconstruction burden by handling unchanging regions, so the deformable set focuses on dynamic parts (\cref{fig:static}).

\subsection{Implementation Details}
\label{sec:implementation}

\paragraph{Positional \& Temporal Encoding:} We facilitate high-frequency deformation fields through PE encoding both the position $\vect{x}$ and time $t$ inputs by $\gamma$, where, for example, $L_{\vect{x}}$ is the respective encoding base for $\vect{x}$:
\begin{equation}
\begin{aligned}
    \gamma(\vect{x}) = (\sin(2^0\vect{x}), \cos(2^0\vect{x}), \sin(2^1\vect{x}), \\ \cos(2^1\vect{x}), \dots, \sin(2^{L_{\vect{x}}-1}\vect{x}), \cos(2^{L_{\vect{x}}-1}\vect{x}))\ .
\end{aligned}
\end{equation}
$L_{\vect{x}}=10, L_t=10$ for both synthetic and real-world scenes. 

\paragraph{Network Architecture:} Our forward-warping DF architecture is inspired by Fang~\etal~\cite{TiNeuVox}'s MLP, with $6$ depth and $256$ width. We use both a time $t$ embedding vector space and a Gaussian position $\mathbf{x}$ embedding vector space.

\paragraph{Optimization:} We use Adam with $\beta=(0.9, 0.999)$ and $\text{eps}=1e^{-15}$. The learning rate for the DF is $0.001$ for all datasets, with exponential scheduling that shrinks to $0.001 \times$ the original learning rate until 30\,K iterations. 
We densify both static and deformable Gaussian sets until 20\,K iterations, and keep optimizing both the Gaussians and the network until 40\,K iterations.


\paragraph{Objective:} We optimize using self-supervised image-based reconstruction loss only given prediction $I$ and groundtruth $I_{gt}$. In early optimization until 20\,K iterations, we use an L2 loss; then, we switch to an L1 loss. This helps to increase reconstruction sharpness late in the optimization while allowing gross errors to be minimized quickly. 
We weighted sum L1/L2 loss with a SSIM loss and weight $\lambda_{ssim}$ to get the final objective at training step $itr$:
\begin{equation}
    L(I, I_{gt}) = 
\begin{cases}
    (1.-\lambda_{ssim})L2(I, I_{gt}) \\+ \lambda_{ssim} (1. - SSIM(I, I_{gt})) & \text{if }  {itr} \leq $2e+4$ \\
     (1.-\lambda_{ssim})L1(I, I_{gt}) \\+ \lambda_{ssim} (1. - SSIM(I, I_{gt})) & \text{else } 
\end{cases}
\label{eq:objective}
\end{equation}

\section{Experiments}

\paragraph{Metrics:} 
\label{sec:metrics}
We measure novel view synthesis performance using PSNR, SSIM~\cite{wang2004image}, MS-SSIM~\cite{wang2003multiscale} and LPIPS~\cite{zhang2018perceptual}. Metric subsets are reported following each dataset's convention. All running times are on one NVIDIA 3090 GPU.

\paragraph{Datasets:}
\label{sec:datasets}
We evaluate all methods on the synthetic \textbf{D-NeRF}~\cite{pumarola2021d} and real-world \textbf{NeRF-DS}~\cite{yan2023nerfds}, \textbf{HyperNeRF}~\cite{park2021hypernerf} datasets. The former consists of 800$\times$800 exocentric 360\degree views of $8$ dynamic objects with large motion and realistic materials. To accommodate all baselines, we train and render all methods at half resolution (400$\times$400) with a white background. The real-world NeRF-DS dataset has monocular dynamic sequences with specular objects captured with a handheld camera.
We also evaluate on the HyperNeRF dataset that contains general dynamic scenes captured by monocular cameras. Some prior methods do not report LPIPS on HyperNeRF; as such, we exclude it. 

\subsection{Results}
\label{sec:comparisons}


\begin{figure*}[t]
  \centering
   \begin{tabular}{p{0.12\linewidth}p{0.12\linewidth}p{0.12\linewidth}p{0.12\linewidth}p{0.12\linewidth}p{0.12\linewidth}p{0.12\linewidth}}
\centering GT & \centering Ours & \centering K-Planes & \centering NDVG & \centering Hexplanes & \centering TiNeuVox & \centering V4D 
\vspace{1mm}
\end{tabular}
   \includegraphics[width=0.99\linewidth]{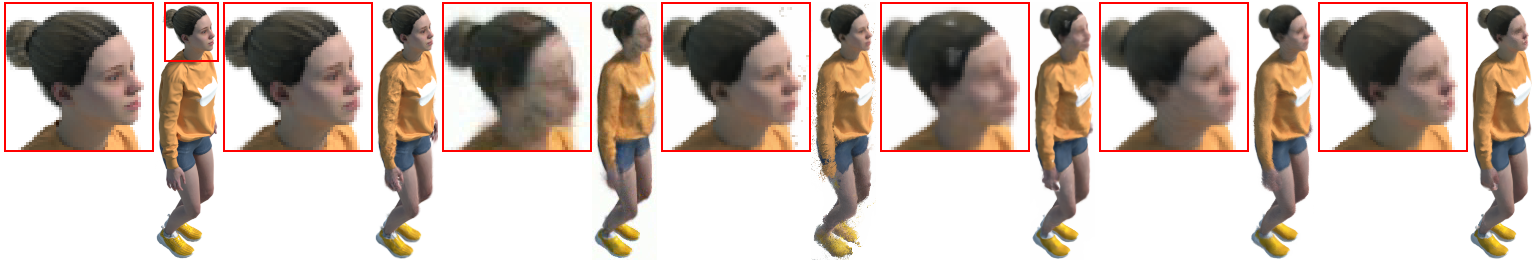}
   \includegraphics[width=1.0\linewidth]{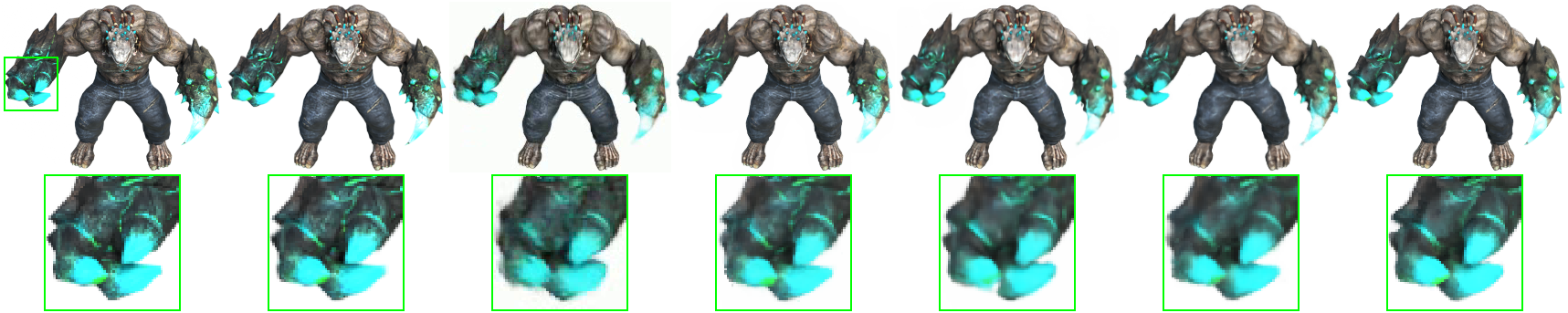}
   \includegraphics[width=1.0\linewidth]{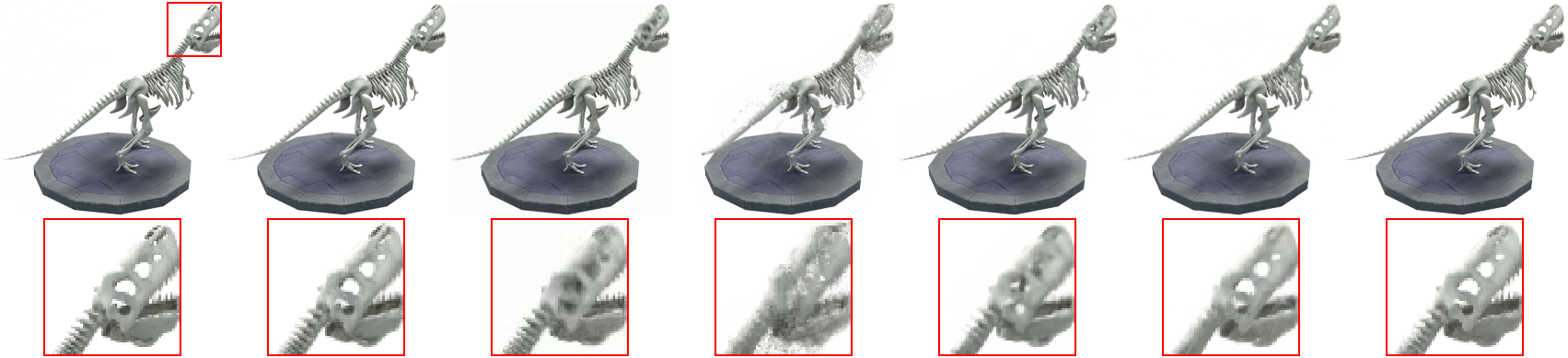}
   \caption{\textbf{Quantitative comparison of GauFRe and the baseline methods for test views from  DNeRF~\cite{pumarola2021d} (400$\times$400) dataset.} All methods reproduce the rough geometry, but sufficient sampling is necessary to reproduce the fine detail. Our approach can efficiently spread Gaussians to both static and dynamic regions to maximize quality, producing the sharpest image of all compared methods.}
   \label{fig:results-qualitative-dnerf}
\vspace{-.5cm}
\end{figure*}
\begin{figure}[t]
    \centering
    \begin{tabular}{p{0.15\linewidth}p{0.15\linewidth}p{0.15\linewidth}p{0.15\linewidth}p{0.15\linewidth}}
        \centering GT & \centering Ours  & \centering NDVG &  \centering TiNeuVox & \centering V4D 
    \end{tabular}
    \includegraphics[width=1.0\linewidth]{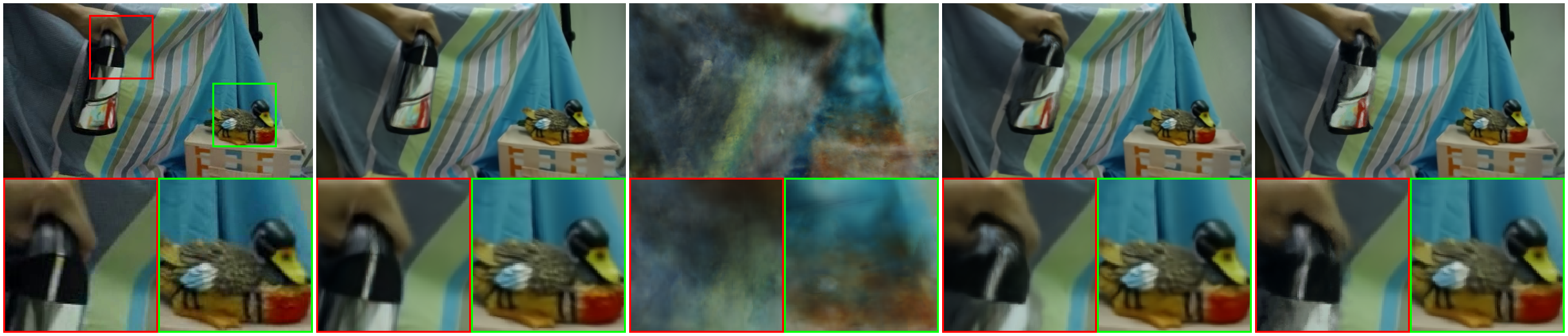}
    \includegraphics[width=1.0\linewidth]{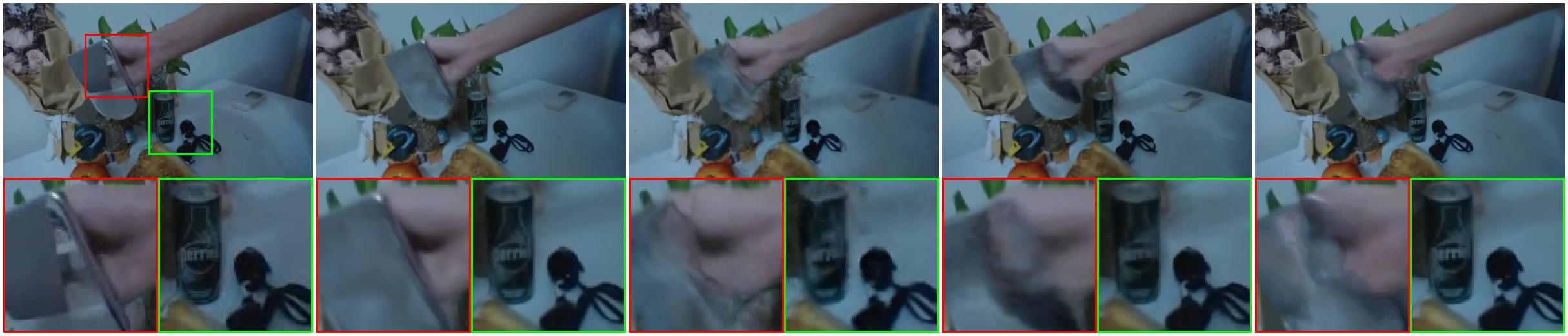}
    \includegraphics[width=1.0\linewidth]{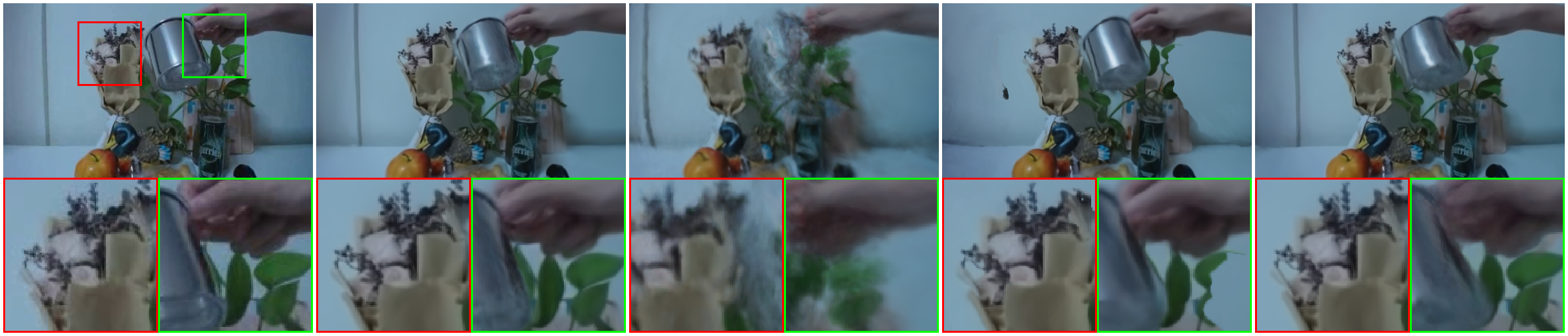}
    \includegraphics[width=1.0\linewidth]{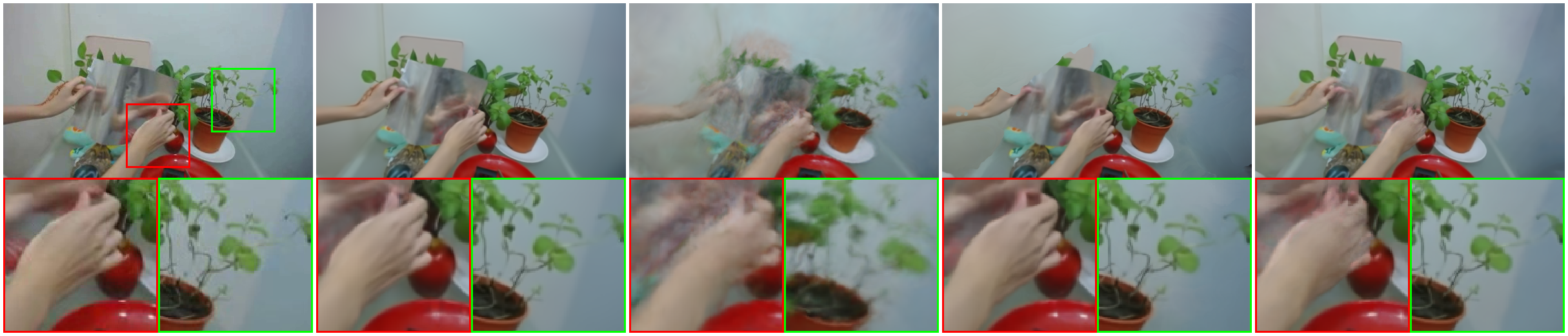}
    \vspace{-7mm}
    \caption{\textbf{Qualitative results on the NeRF-DS~\cite{yan2023nerfds} monocular video dataset.} Compared to the baseline methods, our approach reproduces finer details on dynamic objects such as hands, and shows overall high quality in static regions.}
    \label{fig:NeRF-DS-quali}

    \vspace{2mm}
    \begin{tabular}{p{0.15\linewidth}p{0.15\linewidth}p{0.15\linewidth}p{0.15\linewidth}p{0.15\linewidth}}
        \centering GT & \centering Ours  & \centering 4DGS &  \centering EffG & \centering D3G 
    \end{tabular}
    \includegraphics[width=1.0\linewidth]{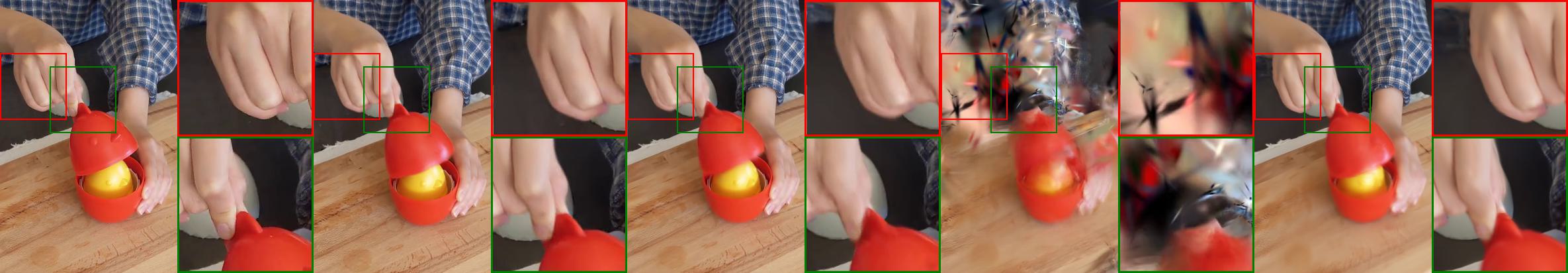}
    \includegraphics[width=1.0\linewidth]{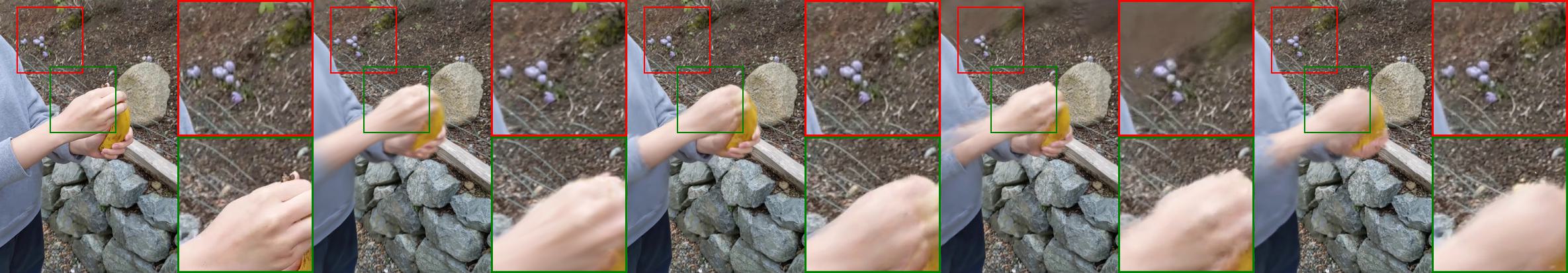}
    \vspace{-7mm}
    \caption{\textbf{Qualitative results on the HyperNeRF dataset for dynamic Gaussian approaches.} Our method achieves comparable if not better results to 4DGS~\cite{wu20234dgaussians}, D3G~\cite{yang2023deformable} and EffG~\cite{katsumata2023efficient}.} 
    \label{fig:HyperNeRF-quali-concurrent}

    \vspace{2mm}
    \begin{tabular}{p{0.15\linewidth}p{0.15\linewidth}p{0.15\linewidth}p{0.15\linewidth}p{0.15\linewidth}}
        \centering GT & \centering Ours  & \centering 4DGS &  \centering EffG & \centering D3G 
    \end{tabular}
    \includegraphics[width=1.0\linewidth]{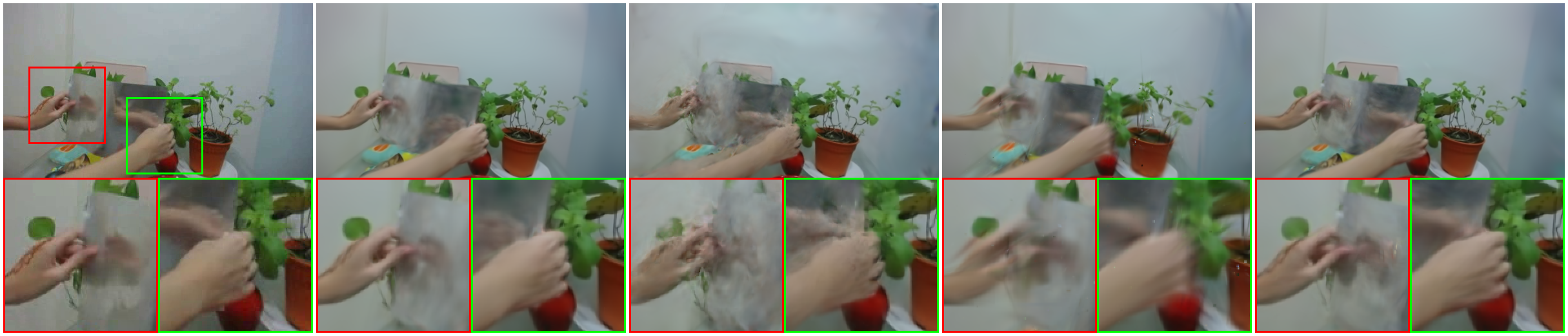}
    \includegraphics[width=1.0\linewidth]{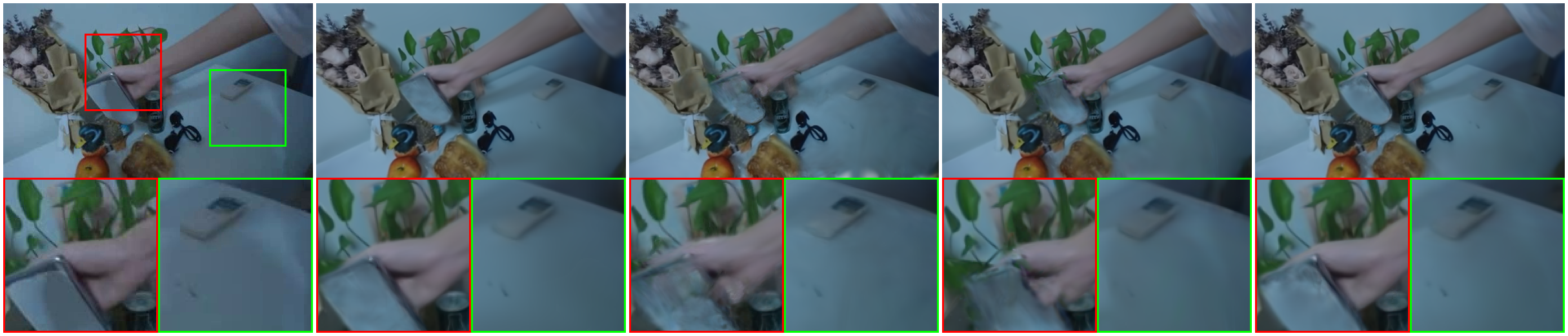}
    \includegraphics[width=1.0\linewidth]{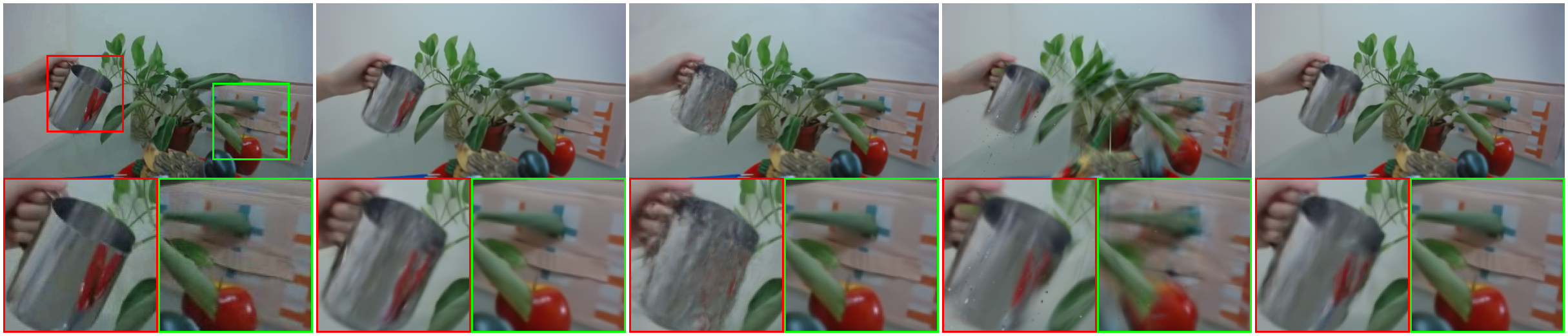}
    \vspace{-7mm}
    \caption{\textbf{Qualitative results on the NeRF-DS dataset for dynamic Gaussian approaches.} Our method achieves comparable results to D3G~\cite{yang2023deformable}, while training and rendering much faster.}
    \label{fig:NeRF-DS-quali-concurrent}
\end{figure}

\paragraph{Deformation Modeling:} We compare with NDVG\cite{guo2022neural} and TiNeuVox\cite{TiNeuVox} on both synthetic and real-world datasets, and with  Nerfies\cite{park2021nerfies}, HyperNeRF\cite{park2021hypernerf} and NeRF-DS\cite{yan2023nerfds} on real-world datasets (\cref{table:results-nvs-quantitative}, \cref{fig:results-qualitative-dnerf}, \cref{fig:NeRF-DS-quali}). Our method achieves high reconstruction quality on both synthetic and real-world datasets, while utilizing the 3D Gaussian's rasterization pipeline to achieve real-time rendering speeds. At the same time, optimizing our method only takes 10--20 minutes on D-NeRF and NeRF-DS, which is considerably faster than the MLP-based NeRF representations.

\paragraph{Efficiency:} Many recent methods have proposed auxiliary structures to accelerate the training and rendering of volume-rendered dynamic NeRFs. Among these, we compare with the voxel-grid variants of NDVG\cite{guo2022neural} and TiNeuVox\cite{TiNeuVox}, and with the plane-based representations of K-Planes\cite{fridovich2023k} and HexPlane\cite{cao2023hexplane}. Compared to the MLP-based Nerfies\cite{park2021nerfies} and HyperNeRF\cite{park2021hypernerf}, these representations trade image quality for efficient rendering. Gan~\etal's V4D\cite{gan2023v4d} seeks to maintain both image quality and rendering speed at the cost of slow training (\cref{fig:results-qualitative-dnerf}, \cref{fig:NeRF-DS-quali}). Our method, on the other hand, achieves high reconstruction quality alongside efficient training and rendering (\cref{table:results-nvs-quantitative}). 

\paragraph{Ablations:}
We ablate the alternatives discussed in \cref{sec:gaussian-deformation-fields} for parameterizing the deformations in (\cref{tab:ablations}). We observe that the ``Fix Scale'', ``Deform Opacity'', and ``Deform SH'' ranked lowest. ``Fix Scale'' corresponds to fixing $\vect{s}_i$, that is, not morphing $\mathcal{G}_i$ with time. Understandably, as most real-world scenes are not strictly rigid this limits the Gaussian's representational power. ``Deform Opacity'' and ``Deform SH'' corresponds to allowing the transparency and appearance to change. Despite increased representational power, these variants strongly overfit the training views leading to lower reconstruction quality. ``Quaternion Addition'' models the deformation on $\vect{q}_i$  by addition instead of the geometrically-correct multiplication. Finally, ``Scale Post-Exponentiate'' confirms that pre-activation is marginally better than post-activation. 

The variants ``No Static Gaussians'' and ``No IB Init'' evaluate the effect of the static separation. The former removes the static primitive set $\{\mathcal{G}_j\}$, allowing all Gaussians to be deformable; the latter replaces our proposed inductive bias-aware initialization with SfM initialization (the default setting for real-world scenes). Both lower performance. Moreover, ``No IB Init'' has worse performance than ``No Static Gaussians'' indicating the necessity of a sound initialization strategy for the static component. In addition, removing L2/L1 loss transit marginally hurt GauFRe performance as shown in ``No LR Transit''.

{
\setlength{\tabcolsep}{2pt}
\begin{table}[t]
\small
\begin{center}
\resizebox{\linewidth}{!}{
    \begin{tabular}{lccccc}
    \toprule
    \multicolumn{6}{c}{D-NeRF~\cite{pumarola2021d} (synthetic)} \\
    \midrule
     & PSNR$\uparrow$ & SSIM$\uparrow$ & LPIPS$\downarrow$ & Optim. $\downarrow$ & Render$\downarrow$\\
    \midrule
    NDVG~\cite{guo2022neural} & 30.5 & 0.97 & 0.054 & 25mins & $>1$s \\
    TiNeuVox~\cite{TiNeuVox} & 32.9 & 0.97 & 0.041 & 20mins & $>1$s \\
    K-planes~\cite{fridovich2023k} & 29.2 & 0.96 & 0.060 & 60mins & $>1$s \\
    Hexplane~\cite{cao2023hexplane} & 31.0 & 0.97 & 0.039 & 15mins & 0.5s \\
    V4D~\cite{gan2023v4d} & 33.4 & 0.98 & 0.027 & 6hours & 0.5s \\
    Ours & 34.5 & 0.98 & 0.023 & 13mins & 0.01s \\
    \midrule
    \multicolumn{6}{c}{NeRF-DS~\cite{yan2023nerfds} (real)} \\
    \midrule
     & PSNR$\uparrow$ & MS-SSIM$\uparrow$ & LPIPS$\downarrow$ & Optim. $\downarrow$ & Render$\downarrow$\\
    \midrule
    Nerfies~\cite{park2021nerfies} & 20.1 & 0.71 & 0.349 & $\sim$hours & $>1$s \\
    HyperNeRF~\cite{park2021hypernerf} & 23.0 & 0.85 & 0.181 & $\sim$hours & $>1$s \\
    NeRF-DS~\cite{yan2023nerfds} & 23.7 & 0.89 & 0.143 & $\sim$hours & $>1$s \\
    NDVG~\cite{guo2022neural} & 19.1 & 0.58 & 0.417 & 1hours & $>1$s \\
    TiNeuVox~\cite{TiNeuVox} & 21.7 & 0.82 & 0.219 & 30mins & $>1$s \\
    V4D~\cite{gan2023v4d} & 23.5 & 0.88 & 0.142 & 7hours & $>1$s \\
    Ours & 23.9 & 0.89 & 0.148 & 20mins & 0.01s \\
    \midrule
    \multicolumn{6}{c}{HyperNeRF~\cite{park2021hypernerf} (real)} \\
    \midrule
     & PSNR$\uparrow$ & MS-SSIM$\uparrow$ & Optim. $\downarrow$ & Render$\downarrow$\\
    \midrule
    Nerfies~\cite{park2021nerfies} & 22.2 & 0.80  & $\sim$hours & $>$ 1s\\
    HyperNeRF~\cite{park2021hypernerf} & 22.3 & 0.81  & $\sim$hours & $>$ 1s\\
    NDVG~\cite{guo2022neural} & 23.3 & 0.82  & 35mins & $>$ 1s \\
    TiNeuVox~\cite{TiNeuVox} & 24.3 & 0.84 & 30mins & $>$ 1s \\
    V4D~\cite{gan2023v4d} & 24.8 & 0.83 & $\sim$hours & $>$ 1s \\
    Ours & 24.1 & 0.83 & 1.5hours& 0.06s\\
    \bottomrule
    \end{tabular}
}
\caption{\textbf{Comparison with deformation-based methods on the dataset from D-NeRF~\cite{pumarola2021d} (400$\times$400), NeRF-DS~\cite{yan2023nerfds} and HyperNeRF~\cite{park2021hypernerf}.} GauFRe demonstrates top image rendering quality, is the fastest to train, and enables real-time rendering.}
\vspace{-.5cm}
\label{table:results-nvs-quantitative}
\end{center}
\end{table}
}

{
\setlength{\tabcolsep}{2pt}
\vspace{-0.3cm}
\begin{table}[t]
\small
\begin{center}
\begin{tabular}{lrrrrr}
\toprule
\multicolumn{6}{c}{NeRF-DS~\cite{yan2023nerfds} (real)} \\
\midrule
 & PSNR$\uparrow$ & MS-SSIM$\uparrow$ & LPIPS$\downarrow$ & Optim. $\downarrow$ & FPS$\uparrow$\\
\midrule
D$^2$NeRF~\cite{wu2022d} & 22.7 & 0.85 & 0.189 & $\sim$hours & $<$1 \\
Ours & 23.9 & 0.89 & 0.148 & 20mins & 96 \\
\midrule
\multicolumn{6}{c}{HyperNeRF~\cite{park2021hypernerf} (real)} \\
\midrule
 & PSNR$\uparrow$ & MS-SSIM$\uparrow$ & LPIPS$\downarrow$ & Optim. $\downarrow$ & FPS$\uparrow$\\
\midrule
NeuralDiff~\cite{tschernezki21neuraldiff} &  19.5 & 0.68 & - & $\sim$hours & $<$1\\
D$^2$NeRF\cite{wu2022d} & 22.1 & 0.80 & - & $\sim$hours & $<$1 \\
Ours & 24.1 & 0.83 & - & 1.5hours & 15\\
\bottomrule
\end{tabular}
\vspace{-2mm}
\caption{\textbf{Comparison with decomposition-aware methods on NeRF-DS~\cite{yan2023nerfds} and HyperNeRF~\cite{park2021hypernerf}.} GauFRe shows better quality while being faster than previous decomposition-aware NeRFs.
}
\label{table:decomposition}
\vspace{-0.25cm}
\end{center}
\vspace{-0.25cm}
\end{table}

}
{
\setlength{\tabcolsep}{2pt}
\begin{table}[t]
\small
\begin{center}
\begin{tabular}{lrrrrr}
\toprule
\multicolumn{6}{c}{D-NeRF~\cite{pumarola2021d} (synthetic)} \\ 
\midrule
 & PSNR$\uparrow$ & MS-SSIM$\uparrow$ & LPIPS$\downarrow$ & Optim. $\downarrow$ & FPS$\uparrow$\\
\midrule
RTGS~\cite{yang2023gs4d} & 34.1 & 0.98 & 0.02 & - & - \\
4DGS~\cite{wu20234dgaussians} & 33.3 & 0.98 & 0.03 & - & - \\
Ours & 34.5 & 0.98 & 0.02 & 13mins & 112 \\
\midrule
\multicolumn{6}{c}{NeRF-DS~\cite{yan2023nerfds} (real)} \\
\midrule
 & PSNR$\uparrow$ & MS-SSIM$\uparrow$ & LPIPS$\downarrow$ & Optim. $\downarrow$ & FPS$\uparrow$\\
\midrule
D3G~\cite{yang2023deformable} & 23.7 & 0.89 & 0.132 & 72mins & 32 \\
4DGS~\cite{wu20234dgaussians} & 22.2 & 0.82 & 0.202 & 98mins & 54 \\
EffG~\cite{katsumata2023efficient} & 21.7 & 0.81 & 0.214 & 8mins & 250 \\
Ours & 23.9 & 0.89 & 0.148 & 20mins & 96 \\
\midrule
\multicolumn{6}{c}{HyperNeRF~\cite{park2021hypernerf} (real)} \\
\midrule
 & PSNR$\uparrow$ & MS-SSIM$\uparrow$ & LPIPS$\downarrow$ & Optim. $\downarrow$ & FPS$\uparrow$\\
\midrule
D3G\cite{yang2023deformable} & 21.4 & 0.68 & - & 2hours & 8\\
4DGS\cite{wu20234dgaussians} & 24.3 & 0.82 & - & 3hours & 20 \\
EffG\cite{katsumata2023efficient} &  21.2 & 0.74 & - & 30mins & 96 \\
Ours & 24.1 & 0.83 & - & 1.5hours & 15 \\
\bottomrule
\end{tabular}
\vspace{-2mm}
\caption{\textbf{Comparison with GS-based methods on D-NeRF\cite{pumarola2021d} (400$\times$400), NeRF-DS~\cite{yan2023nerfds} and HyperNeRF~\cite{park2021hypernerf}.} GauFRe achieves competitive quality, optimization speed, and rendering speed among network-based Dynamic GS representations.
}
\label{table:concurrent}
\vspace{-0.25cm}
\end{center}
\vspace{-0.25cm}
\end{table}
}
{\setlength{\tabcolsep}{1pt}
\begin{table}[]
    \fontsize{9pt}{11pt}\selectfont
    \centering
    \begin{tabular}{l ccc}
    \toprule
    \multicolumn{4}{c}{NeRF-DS~\cite{yan2023nerfds} (real)} \\
    \midrule
    & PSNR$\uparrow$ & MS-SSIM$\uparrow$ & LPIPS$\downarrow$ \\
    \midrule
    Fix Scale  & 20.0 & 0.69 & 0.300 \\   
    Deform Opacity  & 21.9 & 0.79 & 0.230 \\
    Deform SH  & 22.1 & 0.80 & 0.210 \\
    Quaternion Addition & 23.1 & 0.86 & 0.165 \\
    No IB Init       & 23.2 & 0.86 & 0.166 \\
    No Static Gaussians        & 23.5 & 0.88 & 0.154 \\
    Scale Post-exponentiate  & 23.8 & 0.89 & 0.155 \\
    No LR Transit & 23.8 & 0.88 & 0.148 \\
    \midrule
    Full                    & 23.9 & 0.89 & 0.148 \\
    \bottomrule
    \end{tabular}
    \vspace{-2mm}
    \caption{\textbf{Model ablations using the NeRF-DS dataset}, ordered by increasing PSNR. Our full model maximizes all three metrics.}
    \label{tab:ablations}
    \vspace{-0.25cm}
\end{table}
}

\paragraph{Comparison with Gaussian-based Works:} 
We compare with D3G~\cite{yang2023deformable}, 4DGS~\cite{wu20234dgaussians} and EffG~\cite{katsumata2023efficient} on real-world datasets in \cref{table:concurrent} and \cref{fig:NeRF-DS-quali-concurrent}, \cref{fig:HyperNeRF-quali-concurrent}. Yang~\etal's D3G~\cite{yang2023deformable} is most similar to our approach as it used a network-based deformation field to represent motion. Despite using a smaller network, however, our method achieves comparable results thanks to the additional static component. In addition, our method trains almost 4$\times$/2$\times$ faster and renders 3$\times$/2$\times$ faster.  Wu~\etal's 4DGS~\cite{wu20234dgaussians} attempts to accelerate reconstruction by using HexPlane~\cite{cao2023hexplane} to model motion. Compared to plain MLP, HexPlane is more prune to overfitting. Thus, their method lags in reconstruction quality and trains slower on challenging scenario as in NeRF-DS. While more efficient to render than D3G, it is 2$\times$ slower than our method. Finally, Network-Free Gaussian Splatting (EffG)~\cite{katsumata2023efficient} represents a group of works that assume that the motion of Gaussian primitives can be approximated using pre-defined functions like polynomials. This method optimizes function parameters to fit each scene, obviating the use of neural networks and, thus, maximizing efficiency. However, this approach suffers on reconstruction quality in the underconstrained monocular data setting. 

We also compare with GS methods that report quality metrics on D-NeRF~\cite{pumarola2021d} at 400$\times$400: RTGS~\cite{yang2023gs4d} and 4DGS~\cite{wu20234dgaussians}, in ~\cref{table:concurrent}. We achieve higher quality than both.





\paragraph{Comparing to Static/Dynamic Decomposition Methods:} There have been attempts to decompose scenes into static and dynamic parts during reconstruction with NeRFs. We compare with NeuralDiff\cite{tschernezki21neuraldiff} and D$^2$NeRF\cite{wu2022d} on real-world datasets quantitatively (\cref{table:decomposition}) and qualitatively visualizing dynamic masks and static scene renderings following D$^2$NeRF's convention (\cref{tab:reb_d2nerf_quali}) to validate GauFRe's static component design. Both NeuralDiff and D$^2$NeRF produce worse novel view rendering while being significantly slower in both training and rendering.

{
\renewcommand{\tabcolsep}{1pt}
\newcommand{\imgwidth}{0.13\linewidth}
\begin{table}[t]
\centering
\begin{tabular}
{cc|ccc|ccc}
& & \multicolumn{3}{c}{D$^2$NeRF} & \multicolumn{3}{c}{Ours} \\
& Input & Mask & Dyn.~& Static & Mask & Dyn.~& Static \\
\midrule
{\tiny\rotatebox{90}{\hspace{0.15cm}Banana}}
& \includegraphics[width=\imgwidth,trim={18pt 260pt 18pt 200pt},clip]{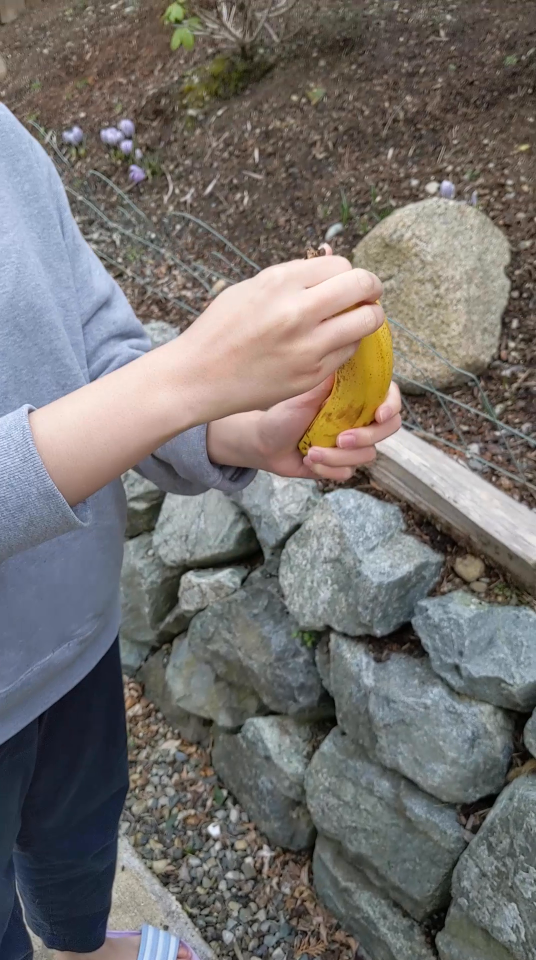}
& \includegraphics[width=\imgwidth]{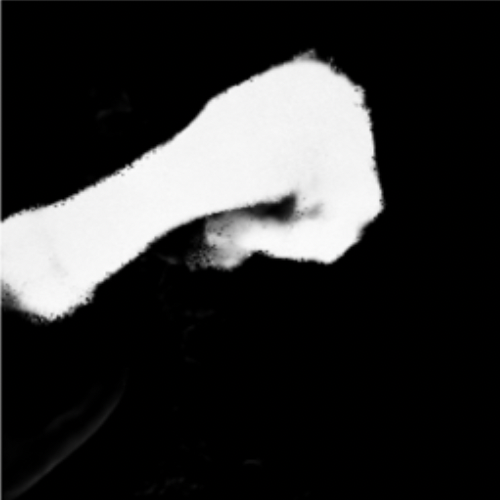}
& \includegraphics[width=\imgwidth]{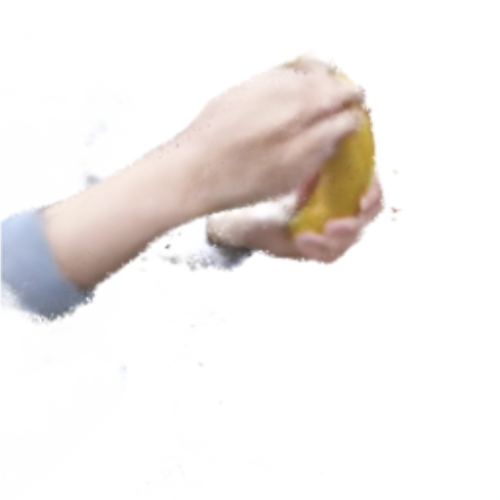}
& \includegraphics[width=\imgwidth]{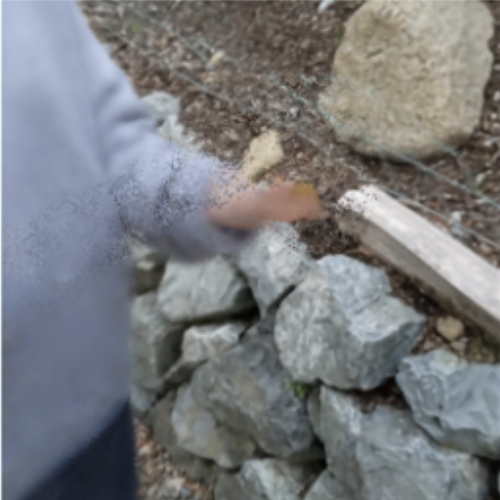}
& \includegraphics[width=\imgwidth,trim={18pt 260pt 18pt 200pt},clip]{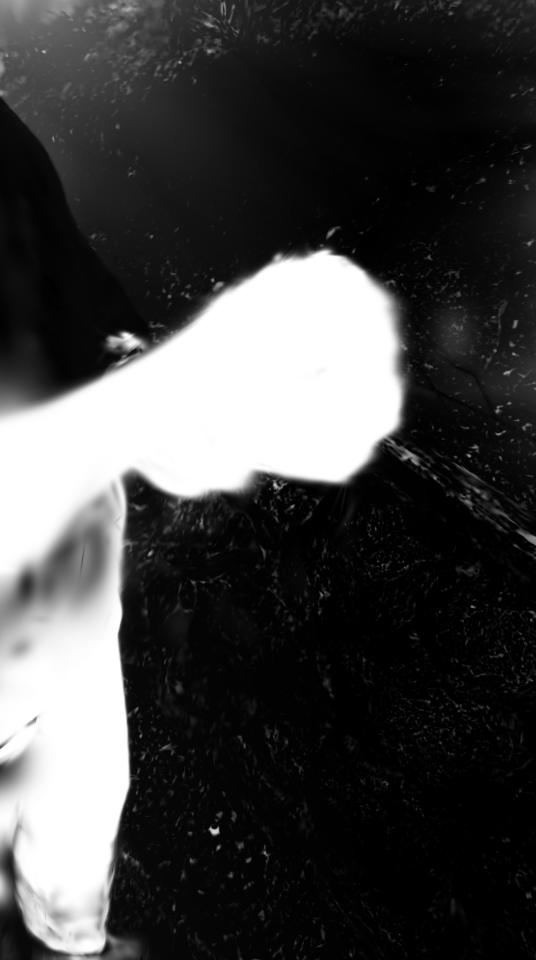}
& \includegraphics[width=\imgwidth,trim={18pt 260pt 18pt 200pt},clip]{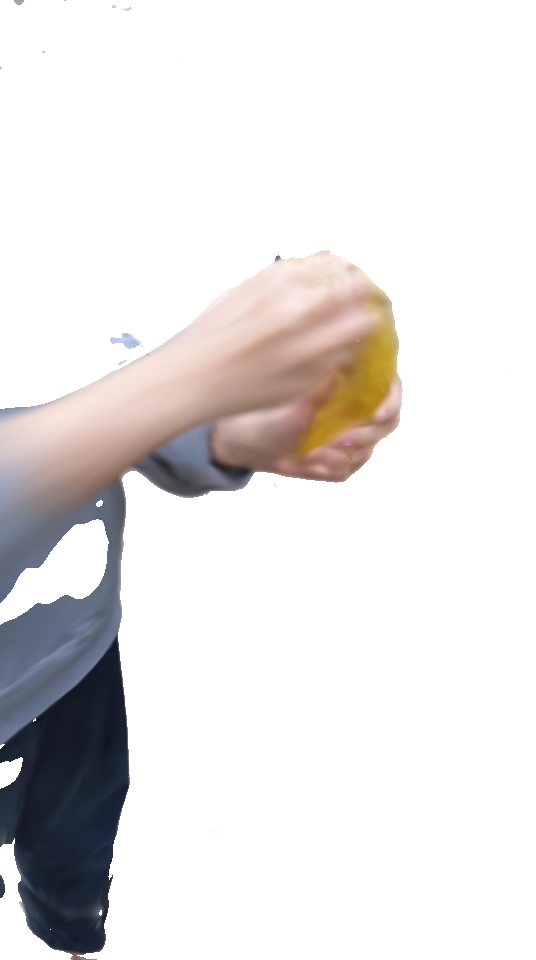} 
& \includegraphics[width=\imgwidth,trim={18pt 260pt 18pt 200pt},clip]{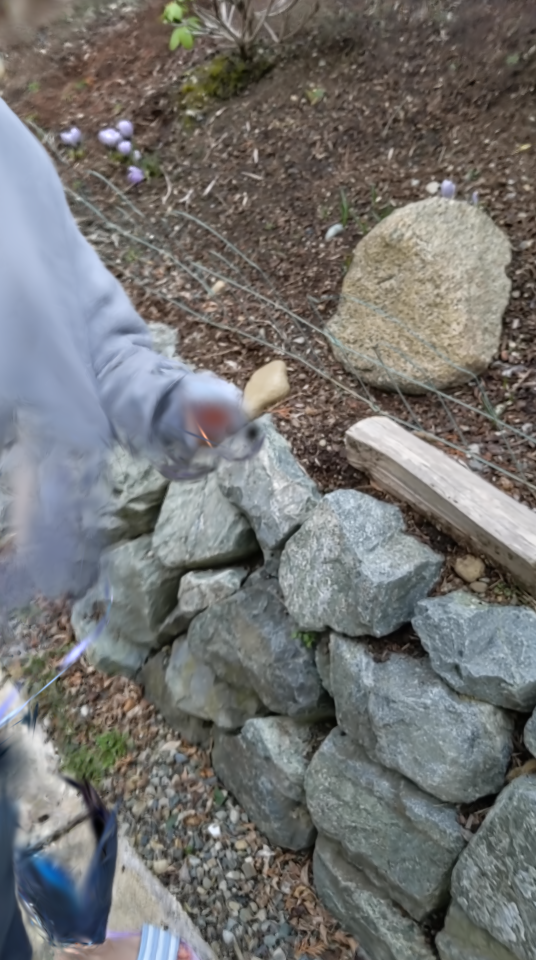} 
\\
{\tiny\rotatebox{90}{\hspace{0.3cm}Bell}}
& \includegraphics[width=\imgwidth,trim={0 20pt 230pt 0},clip]{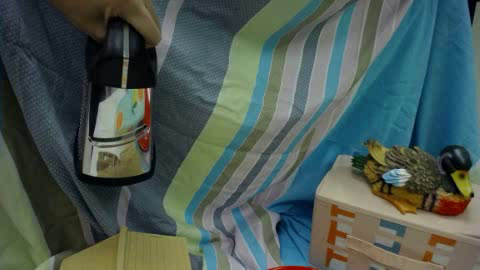}
& \includegraphics[width=\imgwidth,trim={0 20pt 230pt 0},clip]{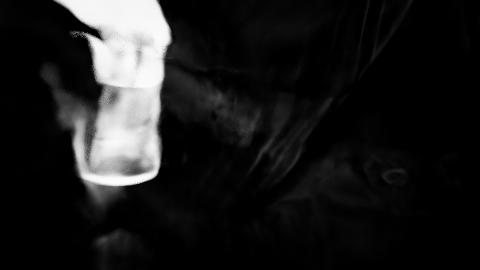}
& \includegraphics[width=\imgwidth,trim={0 20pt 230pt 0},clip]{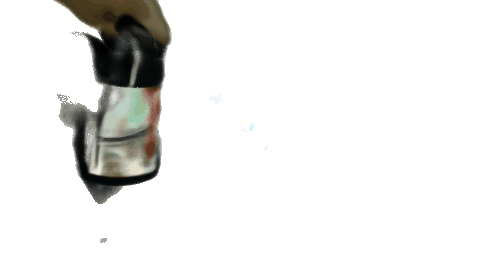}
& \includegraphics[width=\imgwidth,trim={0 20pt 230pt 0},clip]{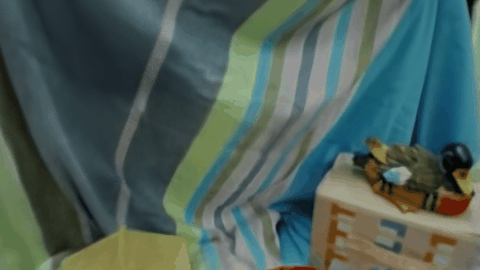}
& \includegraphics[width=\imgwidth,trim={0 20pt 230pt 0},clip]{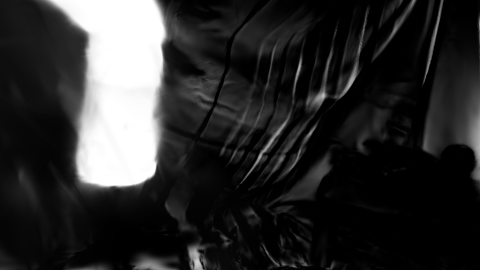}
& \includegraphics[width=\imgwidth,trim={0 20pt 230pt 0},clip]{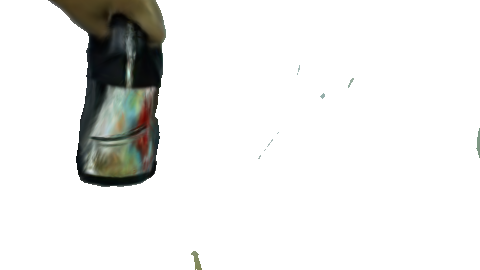} 
& \includegraphics[width=\imgwidth,trim={0 20pt 230pt 0},clip]{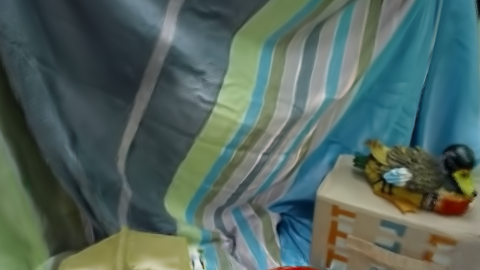} \\
\end{tabular}

\vspace{-0.25cm}
\caption{\textbf{Scene decomposing (qualitative).} \emph{(Top)} HyperNeRF~\cite{park2021hypernerf}; \emph{(Bottom)} NeRF-DS~\cite{yan2023nerfds} dataset. GauFRe's static/dynamic decomposition is comparable to previous state-of-the-art decomposition-aware method.}
\vspace{-5mm}
\label{tab:reb_d2nerf_quali}
\end{table}
}
\section{Conclusion}
In the tricky setting of monocular dynamic scene reconstruction, we must carefully regularize the solution as the problem is underconstrained. GauFRe lets a Gaussian primitive set reside in a canonical space that is affected by an MLP-based deformation field, where the MLP helps to optimize plausible motion solutions. We also use a static scene component and an initialization that guides the deformation field to focus on moving regions of the scene. 
The pipeline is optimized end-to-end with self-supervised image-based rendering loss. 
Experiments show that GauFRe achieves competitive qualitative/quantitative results and efficiency compared to previous state-of-the-art methods, enabling 96\ FPS rendering of real-world scenes on one RTX 3090 GPU. 

\paragraph{Limitations.} GauFRe makes progress on representing dynamic scenes for fast rendering, but it still suffers from issues shared by Gaussian-based methods. Gaussian primitive systems are prone to overfitting: they are powerful enough to recover training views while totally failing to reconstruct the scene, especially when observation is sparse and motion is complex. 
Dynamic GS may struggles with large motions, for which the iterative optimization struggles to correspond primitives; or for thin structures, which are hard to represent accurately with Gaussians.
Concerning densification, one way for dynamic GS methods is to use the original densification policy proposed in 3D-GS~\cite{kerbl2023gaussians}, which sometimes does not cope well with motion.

\paragraph{Acknowledgements.} YL and JT thank a gift from Meta, NSF CNS 2038897, and NSF CAREER 2144956.

\clearpage  

{\small
\bibliographystyle{ieee_fullname}
\bibliography{egbib}
}

\end{document}